\def\year{2018}\relax
\documentclass[letterpaper]{article} 
\usepackage{aaai18}  
\usepackage{times}  
\usepackage{helvet}  
\usepackage{courier}  
\usepackage{url}  
\usepackage{graphicx}  
\usepackage{amssymb,amsthm,amsmath,latexsym}
\theoremstyle{plain}

\newtheorem{conj}{Conjecture}
\title{A Guiding Principle for Causal Decision Problems}
\author{M. Gonzalez-Soto, L.E. Sucar, H.J. Escalante\\
Coordinaci\'on de Ciencias Computacionales,\\
Instituto Nacional de Astrof\'isica \'Optica y Electr\'onica (INAOE)\\
Luis Enrique Erro 1, Santa Maria Tonanzintla, Puebla, M\'exico.
}

\begin{document}
\maketitle
\begin{abstract}
We define a Causal Decision Problem as a Decision Problem where the available actions, the family of uncertain events and the set of outcomes are related through the variables of a Causal Graphical Model $\mathcal{G}$. A solution criteria based on Pearl's Do-Calculus and the Expected Utility criteria for rational preferences is proposed. The implementation of this criteria leads to an on-line decision making procedure that has been shown to have similar performance to classic Reinforcement Learning algorithms while allowing for a causal model of an environment to be learned. Thus, we aim to provide the theoretical guarantees of the usefulness and optimality of a decision making procedure based on causal information.
\end{abstract}

\section{Introduction}
Decision making under uncertainty is a fundamental part of intelligent reasoning \cite{lake2017building} and many real-world applications rely on decisions made by an autonomous agent, such as self-driving cars. Current decision making methods rely on associative methods, which find only statistical patterns in data. On the contrary, causal knowledge allows both for planning and counterfactual reasoning as well as interpretability and explainability \cite{spirtes2000causation}, \cite{woodward2005making}, \cite{pearl2018why}.We propose in this work a way of considering causal information for decision making under uncertainty with rational preferences in such a way that optimal actions are chosen according to the principle of Maximum Expected Utility, which is the formal criteria for making choices under uncertain conditions if rationality is assumed \cite{bernardo2000bayesian}.

\section{Rationality and Expected Utility}
Rationality in a Decision Making setting is defined axiomatically in a way that the \textit{preferences} of a decision maker are logically consistent. If rational preferences are assumed, then it is known that the coherent criteria for making choices is the maximization of expected utility, either with respect to a known utility function and probability distribution or a pair of \textit{subjective} objects \cite{bernardo2000bayesian}, \cite{gilboa2009decision}. Rational decision making has been the standard theory in economics both as a \textit{normative} and a \textit{descriptive} theory of human behavior and it has been the subject of multiple debates \cite{tversky1974judgment}, \cite{kahneman1982judgment}. In this work we pretend to take a \textit{normative} view for a rational decision maker who faces an uncertain environment which is controlled by some unknown causal mechanism.

\section{Optimal Policies}
Given a Reinforcement Learning Problem defined over a Markov Decision Process, a \textit{policy} is a function from the space of \textit{states} to the space of \textit{actions} which is interpreted as what should an agent do in a given state \cite{sutton1998reinforcement}. An \textit{optimal policy} is a policy which is optimal in the sense of achieiving the maximum possible expected reward. Optimal policies can be characterized by the Bellman equations \cite{puterman2005markov}, from where it can be shown to beequivalent to finding the optimal action in the sense of the maximum expected utility \cite{webb2007game}.

\section{Related Work}
Human use and learning of Causal Relations has been extensively studied by Cognitive Scientists. In particular, \cite{hagmayer2009decision}, \cite{wellen2012learning} \cite{hagmayer2013repeated} show that human beings conceive their actions in their environment as interventions over it and they are able to learn, use, and modify previous causal knowledge during a sequential decision process.

From the Machine Learning point of view, \cite{lattimoreNIPS2016} consider a \textit{bandit problem} where the actions available to an autonomous agent are interventions over a known causal model. In their work it is required that the causal model is known, an assumption later relaxed by \cite{sen2017identifying} who considers as unknown only a part of the causal model. 

Our formulation of a Causal Decision Problem attempts to give a framework for an agent to learn optimal actions where a causal model controls his environment and the agent is aware of this. 

\section{Causal Decision Problems}
We define a Causal Decision Problem under Uncertainty (CDPU) as a tuple $(\mathcal{A}, \mathcal{E}, \mathcal{C}, \mathcal{G}, \succeq)$ where  $(\mathcal{A}, \mathcal{E}, \mathcal{C}, \succeq)$ is a classical Decision Problem under Uncertainty \cite{bernardo2000bayesian} and $\mathcal{G}$ is a Causal Graphical Model \cite{sucar2015probabilistic} such that the set of available actions $\mathcal{A}$ and the set of outcomes $\mathcal{C}$ are related through the variables of the Causal Model $\mathcal{G}$; i.e., the events in the family $\mathcal{E}$ correspond to variables in $\mathcal{G}$. It is assumed that the agent does not know the Causal Model $\mathcal{G}$, which is equivalent to not knowing the probabilities of the events $E \in \mathcal{E}$. The model $\mathcal{G}$ is also assumed to remain fixed and to be invariant under interventions \cite{woodward2005making} and to satisfy the conditions expressed in \cite{spirtes2000causation}. The variable in $\mathcal{G}$ which encodes the consequence of the action taken by the agent will be referred as \textit{target variable} since it is the variable where the agent whishes to obtain a desired result. 

In this way, in a CDPU we have a rational agent who chooses an action $a$ among the many available in $\mathcal{A}$, then this action will produce some random effect in the environment which will \textit{cause} a certain consequence, or outcome $c \in \mathcal{C}$. Since 

\section{Proposed Solution}
Since rationality is assumed we must seek to maximize the expected utility of the agent in terms of his current knowledge, which is expressed as a (subjective) probability distribution. Using the awareness of the agent about a causal mechanism governing the environment, intuition encourages to use causal relations to cause some desirable action, as expressed by \cite{joyce1999foundations}

Consider a Causal Decision Problem as stated above where the target variable, call it $Y$, takes its values in the set $\{0,1\}$ and without loss of generality assume that $1$ is the desired output for the agent, then he must choose the action $a^\ast \in \mathcal{A}$ such that
\[ P(Y=1 | do(a^\ast)) \geq P(Y=1 | do(a)) \textrm{ for all } a \in \mathcal{A}. \]
Since the action chosen is the action with the highest probability of producing the most desired action, then it is the action that maximizes the expected utility for the agent. If an action $a_0 \in \mathcal{A}$ yielded a higher expected utility, that would imply that it has higher probability of \textit{causing} the same desired action, but this is not possible because of how $a^\ast$ is obtained.

\section{On-line decision making and causal learning}
In \cite{gonzalez2018playing} a decision-making procedure was proposed using the Proposed Solution together with Bayesian \textit{belief updating} procedure in order to learn an optimal action while acquiring a causal model in the environment and using the current causal knowledge to make choices, this was applied in the simpler case where the decision maker knows the \textit{structure} of the model $\textit{G}$. In the referred work, a decision maker held \textit{beliefs} about the causal information of the environment, which were encoded into probability distributions. Those beliefs were used \textit{as if} they were the true model governing the environment in order to choose an action according to the Solution proposed here. Beliefs were updated in a Bayesian way after observing the causally produced outcome, or consequence, of the action chosen by the agent. 

The actions learned after a series of decision rounds obtained similar performance (in terms of average reward) as an agent learning using the classical Q-Learning procedure. Experimentally, this shows that causal information allows an agent to learn an optimal action, in the sense of expected utility, as well as learning a causal model of the environment.

\section{Experiments}
In \cite{gonzalez2018playing} a test scenario about a medict trying to learn an optimal treatment for a sick patient was used in order to show how causal information could effectively guide a decision making process while also allowing for learning of a causal model. 

We reproduce here the results obtained in \cite{gonzalez2018playing}, which show the average reward of a \textit{causal agent}, where the agent knows the structure of the graphical model and helds beliefs about the \textit{parameters} of the true causal model. The beliefs of the causal agent are used in each decision round as if they were the truth about the causal relations and used as stated above. Comparison against an agent simply choosing at random is also shown.
 
\begin{figure}[ht]
\vskip 0.2in
\begin{center}
\centerline{\includegraphics[width=\columnwidth]{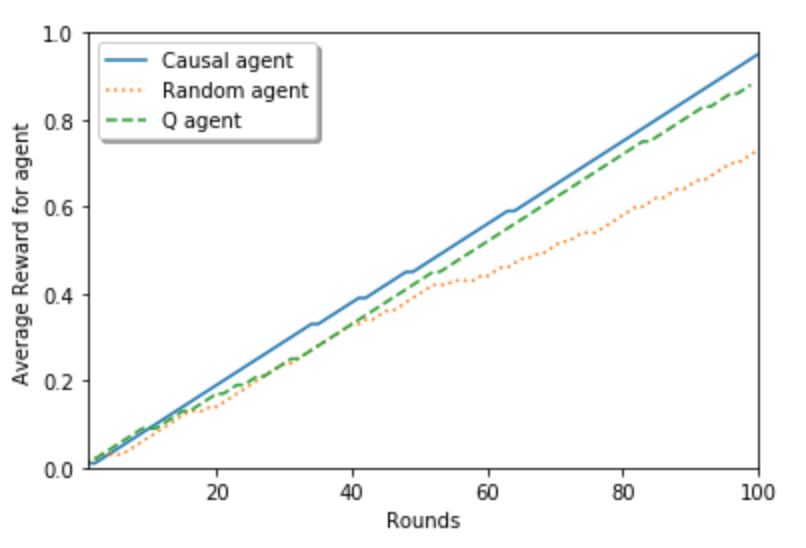}}
\caption{Average reward obtained in each round for each agent}
\label{100_rounds}
\end{center}
\vskip -0.2in
\end{figure}

In Figure \ref{100_rounds} we observe the average reward obtained by the three agents in 100 rounds, where our algorithm slightly outperforms Q-Learning.

\begin{figure}[ht]
\vskip 0.2in
\begin{center}
\centerline{\includegraphics[width=\columnwidth]{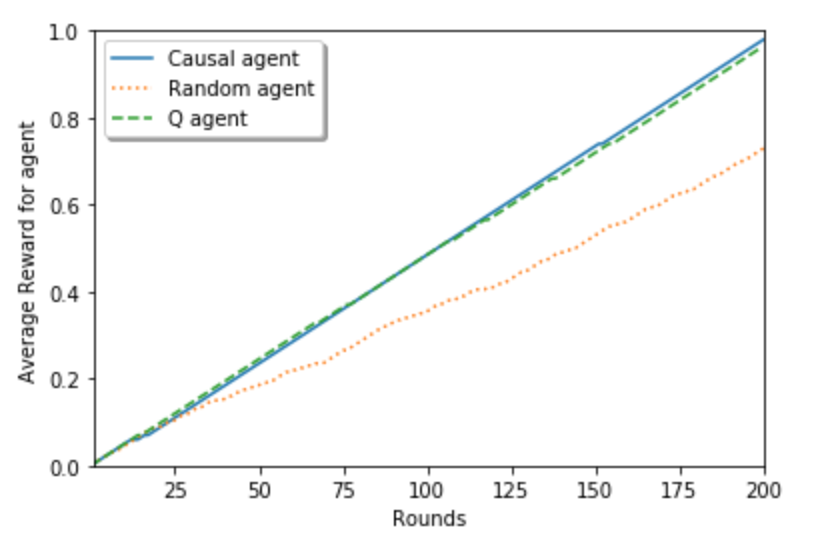}}
\caption{Average reward obtained in each round for each agent}
\label{200_rounds}
\end{center}
\vskip -0.2in
\end{figure}

In Figure \ref{200_rounds} we observe the average reward obtained by the three agents in 200 rounds. The average reward obtained is very similar for Q-learning and our algorithm.

\section{Future Work}
Numerical results show that the trayectories of the average rewards both from our Causal Agent and the Q-agent seem to stay together after some number of rounds, while leaving behind the random-agent. In order to get a valid form of concurrent validation it is required that this behavior will remain like that from a certain point. \\
We thus state a conjecture which is yet to be proven:
\begin{conj}
Let $(X_1,X_2,...) \in \mathbb{R}^\infty$ the rewards obtained by a decision-making procedure which is known to converge to the max expected utility (or an optimal policy), then, if $(Y_1, Y_2,...) \in \mathbb{R}\infty$ are the rewards obtained by a decision-making procedure that uses causal information in the way we propose, then for all $\varepsilon > 0$ there exists an $N_\varepsilon \in \mathbb{N}$ such that $ | X_t - Y_t | < \varepsilon$ for any $t > N_\varepsilon$. 
\end{conj}

\section{Conclusions}
We have proposed an optimality criterion for decision making under uncertainty when the environment where the agent is situated is governed by an unknown causal mechanism. This criteria, when used to build a decision making procedure yields similar performance as classical algorithms which aim towards the same objective: maximizing expected utility, thus showing that causal information, and causal-based decision making is at least as useful as associative-based case, while it also allows to learn a causal model of the environment. The experiments shown serve as a form of concurrent validation, where our proposed method is compared to a decision-making method that it is known to learn optimal policies (i.e., maximize expected utility)

Learning a causal model is useful because of the interpretability and explainability that it provides when analyzing \textit{why} a particular decision was made. For the hypothetical scenario used in the experiments, the causal model allows for further inquiries about the choices made by the medic. As \cite{pearl2018why} mention, the three levels of causal learning are observing, intervening and counterfactual reasoning. Our proposed decision-making method based on the solution stated above allows for each of the three levels to be used. First of all, it allows to observe (and learn) of effects of interventions. In second place, it allow to intervene, and in third place to have the ability of explaining \textit{why} a particular choice was made in terms of the effects it would produce given a certain level of causal knowledge.

It is left as future work to provide a decision making procedure when the causal model is completely unknown for the agent and to provide theoretical convergence results both to the true causal model and the optimal action.
\bibliography{Bibliografia.bib}
\bibliographystyle{aaai}
\end{document}